\documentclass[twoside,twocolumn]{article}

\usepackage{blindtext} 

\usepackage[sc]{mathpazo} 
\usepackage[scale=0.93]{tgpagella} 
\usepackage[T1]{fontenc} 
\usepackage{microtype} 

\usepackage[english]{babel} 

\usepackage[hmarginratio=1:1,top=32mm,columnsep=20pt]{geometry} 
\usepackage[hang, small,labelfont=bf,up,textfont=it,up]{caption} 
\usepackage{float}
\usepackage{booktabs} 

\usepackage{lettrine} 

\usepackage{enumitem} 
\setlist[itemize]{noitemsep} 

\usepackage{abstract} 

\usepackage{titlesec} 
\renewcommand\thesection{\Roman{section}} 
\titleformat{\section}[block]{\large\scshape\centering}{\thesection.}{1em}{} 
\titleformat{\subsection}[block]{}{\thesubsection.}{1em}{} 

\usepackage{fancyhdr} 
\usepackage{titling} 

\usepackage[dvipsnames]{xcolor}
\usepackage{hyperref}

\colorlet{mylinkcolor}{Black}
\colorlet{mycitecolor}{Black}
\colorlet{myurlcolor}{Blue}
\usepackage{graphicx}
\usepackage{cite}

\hypersetup{
  linkcolor  = mylinkcolor,
  citecolor  = black,
  urlcolor   = myurlcolor,
  colorlinks = true,
}

\usepackage{graphicx}

\newcommand{\gopalORCID}{0000-0002-9413-6202}
\newcommand{\nickORCID}{0000-0002-8037-5843}
\newcommand{\adamORCID}{0000-0002-3102-7623}

\pretitle{\begin{center}\huge\bfseries} 
\posttitle{\end{center}} 
\title{AI Safety and Reproducibility: Establishing Robust Foundations for the Neuropsychology of Human Values} 
\author{
\textsc{
Gopal P. Sarma\href{http://orcid.org/\gopalORCID}{\includegraphics[scale=.10]{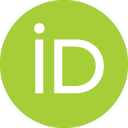}}\textsuperscript{1}\thanks{Email: gopal.sarma@emory.edu},\hspace{2pt} 
Nick J. Hay\href{http://orcid.org/\nickORCID}{\includegraphics[scale=.10]{orcid128}}\textsuperscript{2}\thanks{Email: nnickhay@gmail.com} \thanks{The views expressed herein are those of the author and 
do not necessarily reflect the views of Vicarious AI.},\hspace{2pt} and 
Adam Safron\href{http://orcid.org/\adamORCID}{\includegraphics[scale=.10]{orcid128}}\textsuperscript{3}\thanks{Email: adamsafron@u.northwestern.edu}} \vspace{8pt} \\ 
\normalsize 1. \emph{School of Medicine, Emory University, Atlanta, GA USA}\\ 
\normalsize 2. \emph{Vicarious AI, San Francisco, CA  USA} \\ 
\normalsize 3. \emph{Department of Psychology, Northwestern University, Evanston IL USA}\\ 
}
\date{} 
\renewcommand{\maketitlehookd}{%
\begin{abstract}
\noindent We propose the creation of a systematic effort to identify and replicate key findings in neuropsychology and allied fields
related to understanding human values.  Our aim is to ensure that research underpinning the value alignment problem
of artificial intelligence has been sufficiently validated to play a role in the design of AI systems.  
\end{abstract}
}

\begin{document}

\maketitle

\section{Anthropomorphic Design of Superintelligent AI Systems} \label{introduction}
There has been considerable discussion in recent years about the consequences of achieving human-level 
artificial intelligence.  
In a survey of top researchers in computer science, an aggregate forecast of 352 scientists
assigned a 50\% probability of human-level machine intelligence being realized 
within 45 years.  In the same survey, 48\% responded that greater emphasis 
should be placed on minimizing the societal risks of AI, an emerging area of study known as ``AI safety''
 \cite{2017arXiv170508807G}. \\

A distinct area of research within AI safety concerns software systems
whose capacities substantially exceed that of human beings along every dimension, that is, superintelligence \cite{bostrom2014superintelligence}.  
Within the framework of superintelligence theory, a core research topic known as the \emph{value alignment problem} is to specify
a goal structure for autonomous agents compatible with human values.  The logic behind the framing of this problem is the following:
Current software and AI systems are brittle and primitive, showing little capacity for generalized intelligence.  However, 
ongoing research advances suggest that
future systems may someday show fluid intelligence, creativity, and true thinking capacity.  Defining the parameters of 
goal-directed behavior will be a necessary component of designing such systems.  Because of the complex and intricate nature
of human behavior and values, an emerging train of thought in the AI safety
community is that such a goal structure will have to be inferred by software systems themselves, rather than pre-programmed
by their human designers.  Russell summarizes the notion of indirect inference of human values by stating three principles that 
should guide the development of AI systems \cite{russell2016should}: 
\begin{enumerate}
\item The machine's purpose must be to maximize the realization of human values. In particular, it has 
no purpose of its own and no innate desire to protect itself.
\item The machine must be initially uncertain about what those human values are.  The machine may 
learn more about human values as it goes along, but it may never achieve complete certainty.
\item The machine must be able to learn about human values by observing the choices that we humans 
make.
\end{enumerate}

In other words, rather than have a detailed ethical taxonomy programmed into them,
AI systems should infer human values by observing and emulating 
our behavior \cite{evans2015learning, evanslearning, russell2016should}.  
The value alignment perspective on building safe, superintelligent agents is a natural 
extension of a broader set of questions related to the moral status of artificial intelligence and issues 
related to the architectural transparency and intelligibility of such software-based agents.  
Many of these questions
are important for systems whose capabilities fall well short of superintelligence, but which can
nonetheless have significant impact on the world.  For instance, medical diagnostic systems
which arrive at highly unusual and difficult to interpret diagnostic plans may ultimately do great harm
if patients do not respond the way the AI system had predicted.  In the medical setting, intelligible AI systems
can ensure that healthcare workers are not subsequently forced to reason about circumstances that 
would not have ordinarily arisen via human diagnostics.  Many researchers believe that similar situations 
will arise in industries ranging from transportation, to insurance, to cybersecurity \cite{bryson, sullins, floridi, amodei2016concrete}.  \\

A significant tension that has arisen in the AI safety community is between those researchers concerned
with near-term safety concerns and those more oriented towards longer-term, superintelligence-related concerns \cite{baum2017reconciliation}.
Are these two sets of issues fundamentally in opposition to one another?  Does researching safety
issues arising from superintelligence necessarily entail disregarding more contemporary concerns?  
Our firm belief is that the answer to this question is ``no."  We are of the viewpoint that there is an organic
continuum extending between contemporary and long-term AI safety issues and that individuals and research
groups can freely pursue both sets of issues without tension.  One of the purposes of this article is to argue
that not only can research related to superintelligence be grounded in contemporary concerns, but moreover,
that there is a wealth of existing work across a wide variety of fields that is of direct relevance to superintelligence.
This perspective should be reassuring to researchers who are either skeptical of or have yet to form an opinion 
on the intellectual validity of long-term issues in AI safety.  As we see it, there is no shortage of concrete
research problems that can be pursued within a familiar academic setting. \\

To give a specific instance of this viewpoint, in a recent article, we argued that ideas from affective neuroscience and related fields may play a key role in 
developing AI systems that can acquire human values.  The broader context of this proposal is an inverse
reinforcement learning (IRL) type paradigm in which an AI system infers the underlying utility function of 
an agent by observing its behavior.  Our perspective is that a neuropsychological understanding
of human values may play a role in characterizing the initially uncertain structure that the AI system refines over time.
Having a more accurate initial goal structure may allow an agent to learn from fewer examples.  For a system
that is actively taking actions and having an impact on the world, a more efficient learning process can directly
translate into a lower risk of adverse outcomes.  Moreover, systems built with human-inspired architectures
may help to address issues of transparency and intelligibility that we cited earlier \cite{bryson, floridi}, but in the novel
context of superintelligence.
As an example, we suggested
that human values could be schematically and informally decomposed into three components: \emph{1) mammalian values, 2)
human cognition, and 3) several millennia of human social and cultural evolution} \cite{mammals}.  
This decomposition is simply one possible framing of the problem.  
There are major controversies within these fields and many avenues to approach
the question of how neuroscience and cognitive psychology can inform the design of future AI systems \cite{sotala2016defining}.  
We refer to this broader perspective, i.e. building AI systems which possess structural commonalities with the 
human mind, as \emph{anthropomorphic design}.  

\section{Formal Models of Human Values and the Reproducibility Crisis}
The connection between value alignment and research in the biological and social sciences intertwines this work with 
another major topic in contemporary scientific discussion, the reproducibility crisis.  Systematic studies conducted recently
have uncovered astonishingly low rates of reproducibility in several areas of scientific inquiry \cite{munafo2017manifesto, Horton2015, Campbell2015a}.  
Although we do not know what the ``reproducibility distribution'' looks like for the entirety of science, the shared incentive
structures of academia suggest that we should view all research with some amount of skepticism.\\

How then do we prioritize research to be the focus of targeted replication efforts?  Surely all results do not merit the same
level of scrutiny.  Moreover, all areas likely have ``linchpin results,'' which if verified, will increase
researchers' confidence substantially in entire bodies of knowledge.  Therefore, a challenge for modern science  
is to efficiently identify areas of research and corresponding linchpin results that merit targeted replication efforts \cite{sarma_replication}.  
A natural strategy to pursue is to focus such efforts around major scientific themes or research agendas.  
The Reproducibility Projects of the Center for Open Science, for example, are targeted initiatives aimed replicating 
key results in psychology and cancer biology \cite{cos_psych, cos_biology}.  \\

In a similar spirit, we propose a focused effort aimed at investigating and replicating results which underpin the 
neuropsychology of human values.  Artificial intelligence has already been woven into the fabric of modern society, a trend
that will only increase in scope and pace in the coming decades.  If, as we strongly believe, a neuropsychological understanding of
human values plays a role in the design of future AI systems, it essential that this knowledge base is thoroughly validated.

\section{Discussion and Future Directions}
We have deliberately left this commentary brief and open-ended.  The topic is broad enough that it merits substantial
discussion before proceeding.  In addition to the obvious questions of which subjects and studies should fall under the
umbrella of the reproducibility initiative that we are proposing, it is also worth asking how such an effort will be coordinated.  
Furthermore, this initiative should also be an opportunity to take advantage of novel scientific practices aimed
at improving research quality, such as pre-prints, post-publication peer review, and pre-registration of study design. The specific task of replication is likely only applicable to a subset of results that are relevant
to anthropomorphic design.  There are legitimate scientific disagreements in these fields and many theories and frameworks 
that have yet to achieve consensus.  Therefore, in addition to identifying those studies that are sufficiently concrete and precise 
to be the focus of targeted replication efforts, it is also our aim to identify ``linchpin'' controversies 
that are of high-value to resolve, for example, via special issues in journals, workshops, or more rapid, 
iterated discussion among experts. \\

We make a few remarks about possible starting points. One source of candidate high-value linchpin findings would be those used by frameworks for understanding the nature of emotions. The extent of innate contributions to emotions is hotly debated, with positions ranging from emotions having their origins in conserved evolutionary programs \cite{panksepp1998affective, damasio2012self} to more recent suggestions that emotions are for the most part constructed through social inference \cite{ledoux2016using, barrett2017emotions}. For example, Barrett suggests that the existing affective neuroscience and ethological literature may be based on questionable interpretations of studies of limited generalizability and uncertain reliability of research methods \cite{ekman1972emotion, barrett2017emotions}. A related discipline is
contemplative neuroscience, a field aimed at correlating introspective insights with a neuroscientific
understanding of the brain.  Highly skilled meditators from the Tibetan Buddhist tradition and others have claimed to have
significant insight into human emotions \cite{harrington2006dalai, solms2002brain}, an understanding which is likely relevant to developing a rigorous characterization of human values.  Other frameworks worth considering in depth are models of social-emotional learning based on predictive coding and Bayesian inference \cite{ainley2016bodily}. In these models, uniquely human cognition and affect arises from factors such as extensive early dependency for homeostatic regulation (e.g. fine-CT fibers contributing to analgesia through vagal stimulation \cite{porges2011early, bjornsdotter2011vicarious}). It has been proposed that this dependence leads to models of self that are strongly shaped by the need to predict the minds of others with whom the developing individual interacts. These reciprocal relationships may be the basis for the kind of joint attention and joint intentionality emphasized by Tomasello and others as a basis for uniquely human social cognition \cite{tomasello1999culturalorigins}. \\

In terms of strategies for organizing this literature, we favor an open science or wiki-style approach in which individuals
suggest high-value studies and topics to be the focus of targeted replication efforts.  Knowledgeable researchers
can then debate these proposals in either a structured (such as the RAND Corporation's Delphi protocol \cite{brown1968delphi}) or unstructured format until consensus is achieved
on how best to proceed.  As we have discussed in the previous section, The Center for Open Science has demonstrated
that reproducibility efforts targeting large bodies of literature are achievable with modest resources \cite{cos_psych, cos_biology}.  \\

Our overarching message:  \emph{From philosophers pursuing fundamental theories of ethics, 
to artists immersed in crafting compelling emotional narratives, to ordinary individuals struggling 
with personal challenges, deep engagement with the nature of human values is a fundamental part of the human experience.  
As AI systems become more powerful and widespread, such an understanding may also prove to be important for ensuring 
the safety of these systems. We propose that enhancing the reliability of our knowledge of 
human values should be a priority for researchers and funding agencies concerned about AI safety and existential risks.}
We hope this brief note brings to light an important set of contemporary scientific issues and we are 
eager to collaborate with other researchers in order to take informed next steps.  \\

\section*{Acknowledgements}
We would like to thank Owain Evans and several anonymous reviewers for insightful discussions on the topics of value alignment and reproducibility in psychology and neuroscience.   

\section*{ORCID}
\makebox[2.5cm]{Gopal P. Sarma} \raisebox{-.26\height}{\includegraphics[scale=.10]{orcid128}} \href{http://orcid.org/\gopalORCID}{\gopalORCID}\\
\makebox[2.5cm]{Nick J. Hay} \raisebox{-.26\height}{\includegraphics[scale=.10]{orcid128}} \href{http://orcid.org/\nickORCID}{\nickORCID}\\
\makebox[2.5cm]{Adam Safron} \raisebox{-.26\height}{\includegraphics[scale=.10]{orcid128}} \href{http://orcid.org/\adamORCID}{\adamORCID}\\

\bibliographystyle{ieeetr}
\bibliography{superintelligence_reproducibility}

\end{document}